\documentclass[journal]{IEEEtran}

\ifCLASSINFOpdf
  \usepackage[pdftex]{graphicx}
  \DeclareGraphicsExtensions{.pdf,.jpeg,.png}
\else
\fi

\usepackage{cite}
\usepackage{amsmath}
\usepackage{amsfonts}
\usepackage{mathtools}

\usepackage{caption}
\usepackage{subcaption}
\usepackage{array}
\usepackage{multirow}
\usepackage{xcolor}
\usepackage[linesnumbered,ruled,vlined]{algorithm2e}

\usepackage{textcomp}
\usepackage{setspace}

\SetCommentSty{mycommfont}
\begin{document}

\title{Crime Prediction with Graph Neural Networks and Multivariate Normal Distributions}
%
%
%

\author{Selim~Furkan~Tekin and 
        Suleyman~S.~Kozat,~\IEEEmembership{Senior Member}
        
\thanks{Selim Furkan Tekin and Suleyman S. Kozat are with the Department
of Electrical and Electronics Engineering, Bilkent University, Bilkent, Ankara 06800,
Turkey e-mail: \{tekin; kozat\}@ee.bilkent.edu.tr
 }
}

\maketitle

\begin{abstract}
Existing approaches to the crime prediction problem are unsuccessful in expressing the details since they assign the probability values to large regions. This paper introduces a new architecture with the graph convolutional networks (GCN) and multivariate Gaussian distributions to perform high-resolution forecasting that applies to any spatiotemporal data. We tackle the sparsity problem in high resolution by leveraging the flexible structure of GCNs and providing a subdivision algorithm. We build our model with Graph Convolutional Gated Recurrent Units (Graph-ConvGRU) to learn spatial, temporal, and categorical relations. In each node of the graph, we learn a multivariate probability distribution from the extracted features of GCNs.  We perform experiments on real-life and synthetic datasets, and our model obtains the best validation and the best test score among the baseline models with significant improvements. We show that our model is not only generative but also precise.
\end{abstract}


%
\IEEEpeerreviewmaketitle

\section{Introduction}
%
%
%
%

\IEEEPARstart{C}{rimes} are unlawful acts endangering public safety, leading to colossal life and economic losses if governments do not take the necessary precautions. Understanding the latent patterns of crimes in a geographical region of a city is highly important for controlling human traffic and preventing upcoming crimes with the required force placement that supervises the police to be "in the right place at the right time." Thus, accurate and reliable forecasting with the most descriptive way is essential to canalize the potential reinforcements.

The crime prediction problem is a complex task that contains spatial, temporal, and categorical complexities with inherent interrelations. Prior works approached the crime prediction problem in two methods. The first method is to divide the city area into a grid, where each cell contains the crime information for each crime category that happened in the past. Thus, the idea is to create a spatiotemporal series with tensors containing numerical data such as crime count for each time step. The second method is to divide the city area into unordered regions and create a graph structure that captures the correlations among the regions with corresponding edge weights and node features. After learning the graph structure, the model forecasts the crime occurrences for all regions for the next time step.

However, the regions in both of the methods correspond to extremely wide sections of the cities. To the best of our knowledge, the minimum size of a region in grid-based methods has 4$\mathrm{km}^2$ area. And, for the graph-based approaches, the regions correspond to the districts of a city, where each district has different sizes. For instance, the districts of Chicago city have a range of size that changes from 2$\mathrm{km}^2$ to 35$\mathrm{km}^2$. Thus, assigning a probability value to such a broad area is not expedient and decreases the useability in real-life situations. 

To this end, one can increase the number of regions to enhance the detailing. However, this leads to problems in two approaches. For the grid-based approach, as the grid size increases, the complexity exponentially grows. The model used for capturing the spatial-temporal relations becomes too complex with the growing number of parameters. Furthermore, the interrelations in neighbors of the grid weaken with the high number of cells that contain no crime events. As the cells with zero crime count take the majority, the sparsity of the grid grows, and the data becomes imbalanced. Consequently, grid-based approaches fail to learn the underlying pattern and perform poorly caused by the memorization. On the other hand,  it is difficult to define a rule to add new regions to the graph in graph-based methods. In addition, as we increase the number of nodes in the graph to be the same as the number of regions in graph-based approaches, we will create a model with the same complexity as the grid-based model. Thus, the problems we meet will be analogous to the issues of the grid-based model.

This paper introduces a probabilistic graph-based model called \textit{High-Resolution Crime Forecaster} (HRCF) that performs grid-based crime prediction to remedy these problems. We learn a multivariate probability distribution instead of assigning one probability value to each node in a graph. We build our graph on spatiotemporal data using a divide-and-conquer algorithm to up weight the minority class. Specifically, the algorithm divides the city area into four regions recursively until each has a total crime count below a threshold value. Next we build our graph, where the nodes of the graph represent the centers of each region,  edge weights represent the distances between two neighbor nodes. We model the underlying dynamics of each region with a multivariate gaussian distribution with the objective of minimizing the cross entropy. The architecture contains graph convolutional neural networks as memory units, which capture the spatial and temporal correlations. Each time step, the model predicts multivariate gaussian distributions for all regions conditioned on the previous observations, allowing us to generate predictions for any resolutions. We demonstrate significant performance gains through an extensive set of experiments compared to the conventional methods and show that our model is not only generative but also more precise.

\subsection{Prior Art and Comparisons}
Numerous studies \cite{ehrlich1975relation, braithwaite1989crime, patterson1991poverty, freeman1999economics, Toole2011, wang2012automatic, gerber2014predicting, Wang2016} analyzed crime data in cities, focusing on potential correlations between various features. However, our study focuses on the algorithmic perspective to increase the model performance in high-resolution forecasting for crime events and locations. Thus, we focused on the works related to our methodology consisting of graphical and grid-based approaches with deep learning models. 

Earlier works, \cite{Bowers2004, Chainey2008}, focused on implementing the Kernel Density Estimation (KDE) in the spatial domain to predict the hotspots on the distribution of crime events. However, performing predictions according to the fitted model is problematic due to the temporal changes in short-term and long-term forecasts. Inspiring from earthquake prediction, the authors of \cite{Mohler2011} implemented a self-exciting point process to exhibit the space-time triggering function and compared their methods with KDE modeled crime hotspot maps. Following this work, the authors introduced a marked point process with the parameterized triggering kernel function and compared their approaches with multivariate Hawkes processes \cite{Mohler2014}. However, modeling with self-exciting processes makes strong assumptions on the data, which decreases the expressive power of the respective processes. Thus, the authors of \cite{Du2016} proposed Recurrent Marked Temporal Point Processes (RMTPP) to model the crime events where recurrent neural network parameterizes the intensity function. RMTPP can learn latent temporal dynamics showed high performance compared to baseline processes including Hawkes and Poisson. 

Following these statistical approaches, deep learning structures appear in literature to model crowd movements and showed higher performance. In \cite{Zhang2016}, authors implemented Deep-learning based prediction model for spatiotemporal data (DeepST), where they model the crowd flow with the grid-based forecasting system. Furthermore, \cite{Zhang2017} improved DeepST architecture by modeling the inflow and outflow of crowds in every city region and called their architecture ST-ResNet. Authors of \cite{Wang2017} adapted the ST-ResNet to predict crime distribution over the Los Angeles dataset and achieved high scores. However, they applied spatial-temporal regularization to the data, which increased the computational cost dramatically. Another grid-based deep learning structure is \cite{Huang2019}, where authors designed multi-model units capturing spatial, temporal, and semantic information and producing predictions for each crime type. They leveraged the performance of attention mechanisms and LSTMs while performing predictions in 11x11 grid size. As designated in \cite{Huang2019}, the model performance decreases as the spatial size increases. Likewise, all the other works above produce predictions with small grid sizes.

Due to the high sparsity and complexity of grid-based approaches, graphical models started to emerge in the literature.  With the flexibility and robustness of Graph Neural Networks (GNN) \cite{wu2020comprehensive}, they showed high performance on spatiotemporal domains. For example, \cite{Jain2016} authors introduced SRNN structure to model human-object interactions on image sequences. In crime prediction, \cite{Wang2018} studied the problem of representing spatiotemporal crime data into a graph structure. They formed the graph structure with modeling of independent Hawkes processes in each node and obtained edge weights. Furthermore, they used SRNN like structure with the multi-layered LSTMs on edges and performed predictions for 50 regions in the CHI crime dataset. Another study \cite{Sun2021} implemented Gated Recurrent Network with Diffusion Convolution modules following a Multi-Layer Perceptron (MLP). Their experiments on CHI data built the graph according to districts where the edge values are the distance between nodes. Recently, \cite{Wang2021} introduced a homophily-aware constraint on the loss function, so that neighboring region nodes share similar crime patterns. Nevertheless, these works are non-generative and work on the districts, which causes the problems mentioned in the introduction. 

\subsection{Contributions}
\begin{itemize}
    \item We introduce a subdivision algorithm to create a graph from grid-based spatiotemporal data to reduce sparsity and make predictions with fewer parameters.
    
    \item We present a novel graphical architecture with multivariate Gaussian distributions to jointly train the micro and macro probabilities and model the actual data distribution. 

    \item With the generative structure of our model, we can produce accurate predictions for the high spatial resolution with fewer parameters. 

    \item Through extensive experiments over real and synthetic datasets, we demonstrate that our method brings significant improvements compared to baseline methods.
\end{itemize}

\section{Problem Definition}
We first define various sets that represent information about the data. Next, we define our graph structure and graph function. Following these definitions, we define our objective and how we train our model.

We divide the city region with a grid-based approach into $I\times J$ disjointed cells, which creates the set $M=(m_{1,1}, \dots, m_{i,j}, \dots, m_{I,J})$. Each cell contains the longitude, $x$, and latitude, $y$, information, which we show as $m_{i,j}=(x, y)$. Next, we partition the $M$ set into $N$ disjoint sets according to the algorithm shown in section III to create the region set $R=(r_{1}, \dots, r_{i}, \dots, r_{N})$, where each region contains arbitrary number of cells.
For a given time window T, we also split T as non-overlapping and consequent time slots $T=(t_1, \dots, t_k, \dots, t_K)$. 

For each region $r_i$, we create event matrices $\mathbf{X}_{i, 1:\mathrm{K}}=(\mathbf{x}_{i,1},\dots, \mathbf{x}_{i,k}, \dots, \mathbf{x}_{i,K})\in \mathbb{R}^{L\times K}$ to denote all the crimes occured during the past K slots, where $L$ is total crime categories. In particular, $x_{i,k}^l \in \mathbb{R}$ represents the total event count observed at the region $r_i$ at time slot $k$ for the crime type $l$. Similarly, $\mathbf{X}_{1:N,t} = \mathbf{X}_{t}\in \mathbb{R}^{N\times L}$ denotes the event matrix for all the regions and categories at the time step $t$.

We formulate region graph as $\mathcal{G}=(\mathcal{V}, \mathcal{E}, \mathbf{A})$, where $\mathcal{V}$ represents the set of nodes with $|\mathcal{V}|=N$ where $N$ is the number of regions and $\mathcal{E}$ is the set of undirected edges between region nodes, $\mathbf{A}$ is the weight matrix. Let $v_i \in \mathcal{V}$ to denote a node and $\epsilon_{ij}=(v_i, v_j) \in \mathcal{E}$ to denote an edge. The weight matrix $\mathbf{A}$ has dimension of $N\times N$ with $A_{i,j} = \mathrm{dist}(v_i, v_j)$ if $\epsilon_{i,j}\in\mathcal{E}$ and $A_{i,j} = 0$ if $\epsilon_{i,j}\not\in\mathcal{E}$. Furthermore, we define the graph function, $g(.)$, which takes the event matrices as the input and produces the state vectors, $\mathbf{h}_{i,t} \in \mathbb{R}^{d_h}$, for each graph node, where $d_h$ is the hidden dimension. We formulate the graph function as:
\begin{equation}
    \{\mathbf{X}_{1}, \dots, \mathbf{X}_{K}, \mathcal{G}\}\xrightarrow[]{\text{g(.)}} \{\mathbf{h}_{1,\mathrm{K}+1}, \dots, \mathbf{h}_{\mathrm{N},\mathrm{K}+1}\},
    \label{eq:graph_func}
\end{equation}
where the details of this function shown in section III. 

\begin{figure}[t]
    \centering
    \includegraphics[scale=0.07]{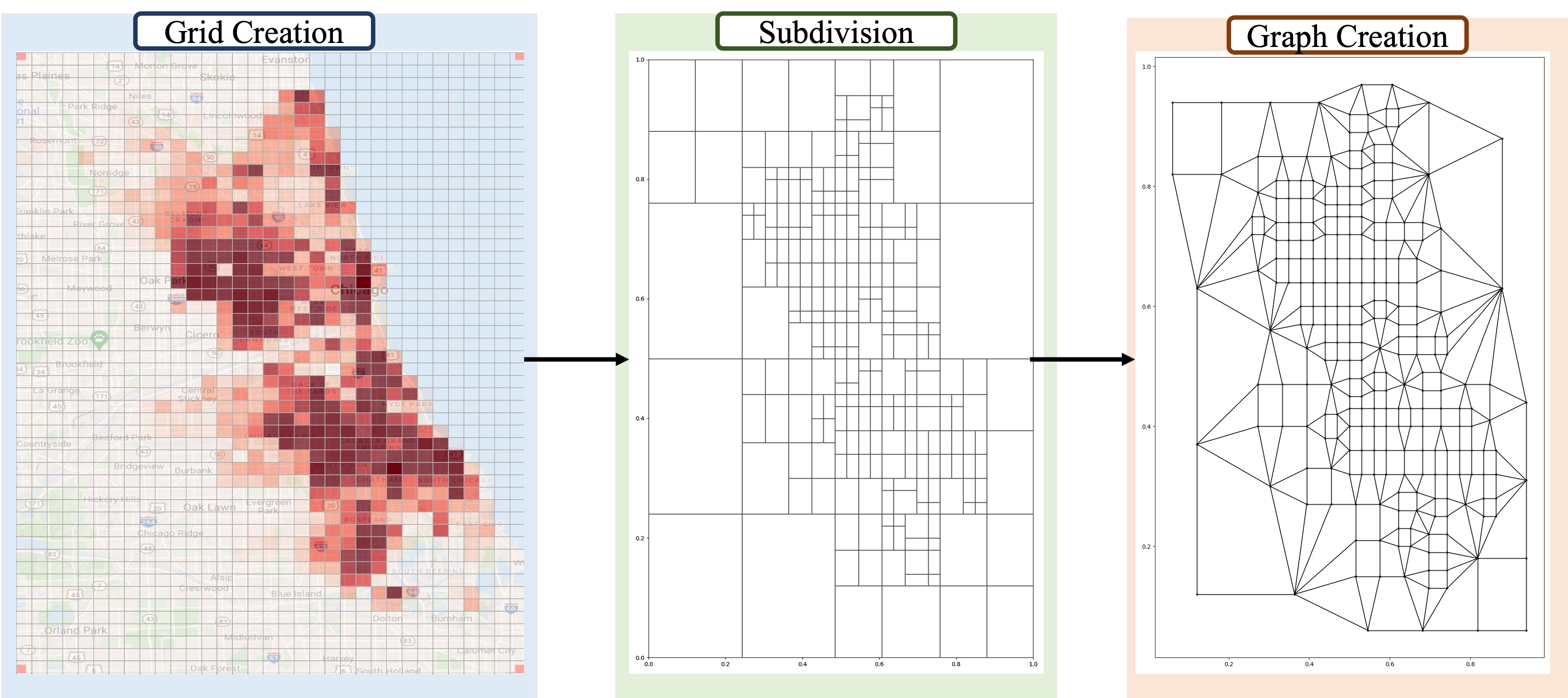}
    \caption{We show the phases of creating our graph. First, we create a grid by dividing the region into $I\times J$ cells, where each cell contains the total event count. Second, with the sampling algorithm that favors the crowded regions, we obtain graph regions. Third, we create the graph with nodes representing the centers of each region and edges representing the distances between nodes.}
    \label{fig:algorithm}
\end{figure}

In this paper, our primary goal is to learn a generative model for crime forecasting using previous crime events. Our first main assumption is that in each time step, $t_k$, the probability of a crime event happening in a region, $r_i$, is governed by a joint probability distribution conditioned on the passed crime events:
\begin{equation}
P(\mathcal{X}_{i,k}, \mathcal{Y}_{i,k} | \mathbf{X}_{1:k-1}) = f_{\mathcal{X}_{i,k},\mathcal{Y}_{i,k}}(x_{i,t}, y_{i,t} | \mathbf{X}_{1:k-1}),
\label{eq:pdf}
\end{equation}
where $x$ and $y$ represent the longitude and latitude of the crime locations. Secondly, we model the joint probability with a likelihood model, $l(x_{i,t}, y_{i,t};\theta(\mathbf{h}_{i,k}))$, and parametrized by the $\theta(.)$ function. Hence, the probability distribution function in equation \ref{eq:pdf} becomes:
\begin{equation}
    f_{\mathcal{X}_{i,k},\mathcal{Y}_{i,k}}(x_{i,t}, y_{i,t} | \mathbf{X}_{1:k-1}) = l(x_{i,t}, y_{i,t};\theta(\mathbf{h}_{i,k})),
\end{equation}
where we decide the likelihood model to be Multivariate Gaussian Distribution. Thus,  $\theta(.)$ parametrize the mean vector and covariance matrix:
\begin{align}
\theta &=(\mathbf{\mu}, \mathbf{\Sigma}) \nonumber, \\
\mathbf{\mu} &= \begin{bmatrix} \mu_{1} & \mu_{2} \end{bmatrix}, \label{eq:mu}\\
\mathbf{\Sigma} &= \begin{bmatrix} v_{1}^2 & 0  \\ 0 & v_{2}^2 \end{bmatrix}, \label{eq:sigma}
\end{align}
where we design the covariates in the covariance matrix to take zero to decrease the number of parameters we learn.
In training, we could directly maximize the log-likelihood (MLE) where we put the locations of each crime event into the likelihood model:
\begin{equation*}
    \mathcal{L} = \sum_{i=1}^{N}\log l(x_{i,t}, y_{i,t};\theta(\mathbf{h}_{i,k})),
\end{equation*}
and sum all the log-likelihoods in each region. However, MLE loss does not directly suffer negative predictions, and in our experiments, we observed that the model stuck on the local minimum. Therefore, we minimize the Kullback-Leibler divergence between the actual distribution and the likelihood model, which is equivalent to minimizing the cross-entropy:
\begin{equation}
    \mathcal{L} = -\sum_{i=1}^{N}\sum_{\substack{x\in\mathcal{X}_{i,t}\\ y\in\mathcal{Y}_{i,t}}} p(x,y)\log l(x, y;\theta(\mathbf{h}_{i,k})),
    \label{eq:loss1}
\end{equation}
where the term $p(x,y)$ is still unknown and note that $x$ and $y$ are continuous values. Recall that we quantized each region with the grid-based approach where each cell takes one if a crime happened, zero otherwise.  With this approach, we can rewrite our equation \ref{eq:loss1} as binary-cross entropy:
\begin{equation}
    \mathcal{L} = -\frac{1}{\mathrm{I}\mathrm{J}}\sum_{i=1}^{N}\sum_{\substack{x\in\mathcal{X}_{i,t}\\ y\in\mathcal{Y}_{i,t}}} q\log l(x, y;\theta) + (1-q)\log(1-l(x,y;\theta)),
    \label{eq:loss2}
\end{equation}
where $q(.)$ maps the longitude and latitude to the crime event value:
\begin{equation}
    q(x, y) = \begin{cases}
        1, & \mathrm{if\,event\,observed\,at}\,(x, y) \\
        0, & \mathrm{otherwise}
   \end{cases}
   \label{eq:idw}
\end{equation}
and we optimize the loss function shown in equation \ref{eq:loss2} with Adam optimizer using batches of training data.

\section{Methodology}
In this section, we describe the subdivision algorithm to create our graph. Next, we present the details of the graph model.

\subsection{Subdivision Algorithm}

\begin{figure}[t]
\centering
    \begin{subfigure}{0.24\textwidth}
        \centering
        \includegraphics[width=\textwidth]{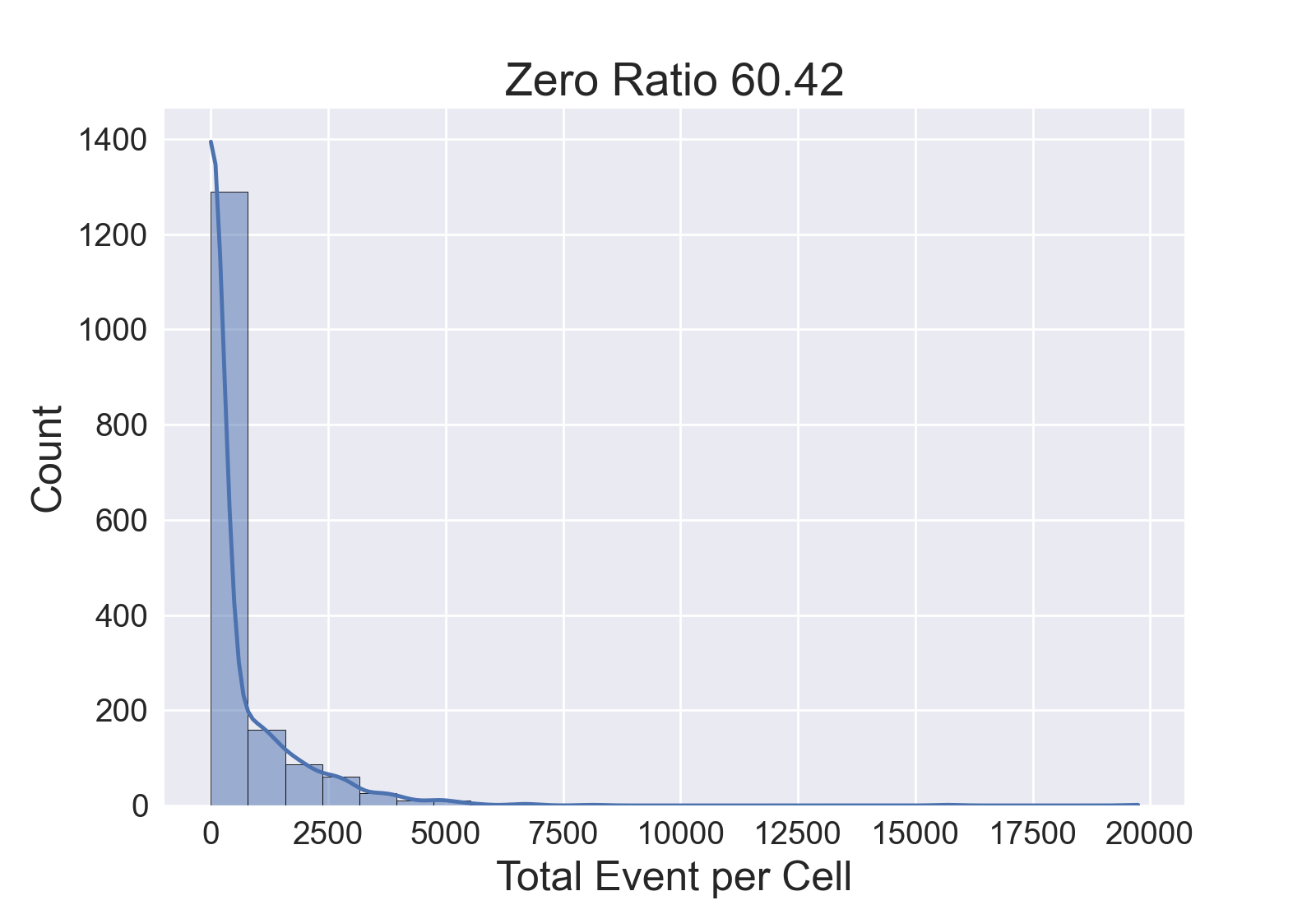}
        \caption{}
        \label{fig:histogram_a}
    \end{subfigure}
    \begin{subfigure}{0.24\textwidth}
        \centering
        \includegraphics[width=\textwidth]{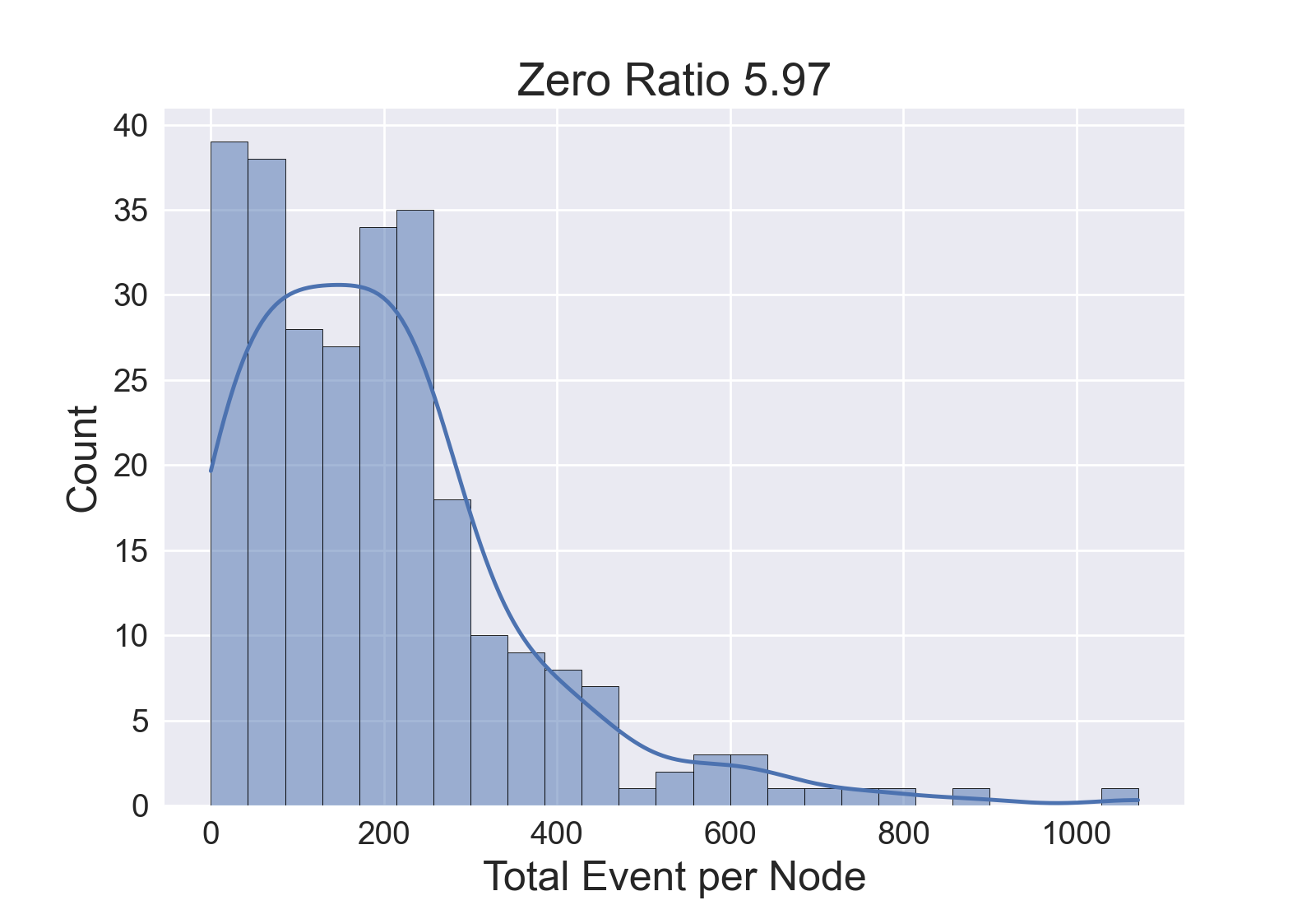}
        \caption{}
        \label{fig:histogram_b}
    \end{subfigure}
    \caption{We show the distribution of events for the grid (a) and the graph (b) structures. While the grid structure contains $60.42\%$ of cells with zero events, the node structure contains $5.97\%$.}
    \label{fig:histogram}
\end{figure}

As we mention in section II, we divide the region into $I\times J$ cells, and each cell contains the information of the longitude and latitude which creates the $M$ set, which is indexed by the $i$th row and $j$th column. Following this definition, we can represent the number of crimes in each cell regardless of crime type at the time $t$ with $\mathbf{E}_{t}=(e_{11},\dots, e_{ij}, \dots, e_{IJ})\in\mathbb{R}^{I\times J}$. Next, we create the grid matrix $\mathbf{Q}\in\mathbb{R}^{I\times J}$ by summing each event matrix at time step $t$:
\begin{equation}
    \mathbf{Q} = \sum_{t=1}^{T}\mathbf{E}_{t},
\end{equation}
where we show the $\mathbf{Q}$ matrix in the first phase of the figure \ref{fig:algorithm}. Observe that the number of cells that contain no events takes the majority. We show this sparsity with the distribution of crime events in each cell in figure \ref{fig:histogram_a}. 

We introduce a subdivision algorithm with a divide-and-conquer approach to overcome this problem, allowing us to use a more flexible graph structure in any spatially distributed data. Algorithm \ref{alg:sampling} shows the subdivision algorithm to create regions. Our goal is to make all the regions contain a total event count less than a threshold value, $\tau$. Our base condition is to have a region larger than $(2, 2)$ or a total event count less than the threshold value. Since the problem contains sub-problems that can be solved individually, we used a recursive approach.

After the subdivision process, we build our graph with the nodes representing the centers of each region and the edges representing the distance between the neighbor nodes. Figure \ref{fig:histogram_b} shows the distribution of events in each region. Note that the change in zero ratios shows the sharp decrease in the sparsity, which can be further decreased according to the threshold and minimum region size selection.

\RestyleAlgo{ruled}
\SetSideCommentLeft
\begin{algorithm}[t]
\SetAlgoLined
\DontPrintSemicolon
\SetNoFillComment
\LinesNotNumbered
\KwIn{$\mathbf{Q}, \tau, r, c$}    
\KwOut{$R$ \tcp*[l]{list of regions}}
    \SetKwFunction{FMain}{DivideRegions}
    \SetKwProg{Fn}{Function}{:}{}
    \Fn{\FMain{$\mathbf{Q}$, $\tau$, $r$, $c$}}{
        \tcc{Select the sub-region}
        $ \mathbf{Q}^{*}  \longleftarrow \mathbf{Q}\mathrm{[r[0]:r[1], c[0]:c[1]]}$

        {\eIf{$(\mathrm{sum}(\mathbf{Q}^{*}) < \tau) \:||\: (\mathbf{Q}^{*}.\mathrm{shape} < (2, 2))$}{
            \textbf{return} [[r, c]] \tcp*[l]{base case}
            }
            {
            I $\gets$ SplitRegions($\mathbf{Q}^{*}$, $r$, $c$) \tcp*[l]{divide 4}
            R $\gets \mathrm{[\:]}$\;
            \ForEach{$r, c \in \mathrm{I}$}{
                R.append(DivideRegions($\mathbf{Q}^*$, $\tau$, $r$, $c$))
                }
            \textbf{return} R\;
            }
        }
}
\textbf{End Function}
\caption{Algorithm for Subdivision}
\label{alg:sampling}
\end{algorithm}

\subsection{High Resolution Crime Forecaster (HRCF)}
\begin{figure*}[t]
    \centering
    \includegraphics[scale=0.1]{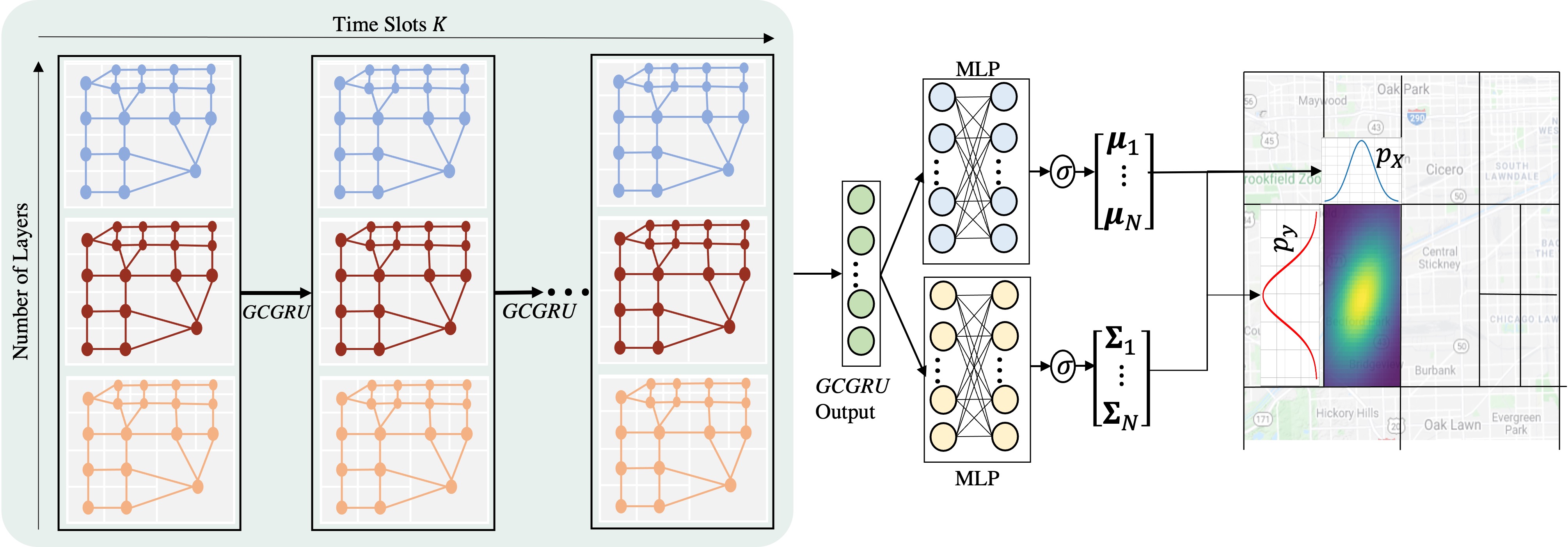}
    \caption{We show HRCF architecture. In each time slot $t_k$, we extract features with the Graph-ConvGRU operations. We pass these features to two different MLP heads, $\mu$ and $\Sigma$. Each head produces the $\mathbf{\mu}_{i}$ and $\mathbf{\Sigma}_{i}$ parameters of distribution in each node.}
    \label{fig:graph_model}
\end{figure*}

This section gives details of each component in our model shown in figure \ref{fig:graph_model} and their contributions to solving our problem. 

A crime event correlates with abnormal events in nearby locations \cite{Wang2016}. An after-effect of a crime can trigger another crime in nearby areas \cite{Mohler2011}. We use graph convolution to learn such relation due to its success of modelling spatial-temporal series\cite{Jain2016}. The idea of graph convolution is analogous to the traditional 2D convolution. An image pixel is similar to a node in a graph, where the filter size of 2D convolution determines the number of neighbors.  The 2D convolution takes the weighted average of pixel values of the central node along with its structured grid neighbors. Similarly, graph convolution takes the weighted average of the central node along with its unordered and variable in size neighbors. Each node collects messages from the input signal to its neighbors. Therefore, we can use convolution operations to carry the information of a crime event in a region to nearby areas.

Mathematically, we motivate the graph convolution based on the steps shown in \cite{wu2020comprehensive}. In graph signal processing Fourier Transform takes the projection of a graph signal $\mathbf{x}\in\mathbb{R}^n$ to orthonormal space of graph, $\mathcal{F}(x) = \mathbf{U}^T\mathbf{x}=\tilde{\mathbf{x}}$, where $\mathbf{U}^T\mathbf{U}=\mathbf{I}$. The inverse transform is also defined as $\mathcal{F}^{-1}(\tilde{\mathbf{x}})=\mathbf{U}\tilde{\mathbf{x}}$. To define the graph convolution an orthonormal basis must be created. Therefore, the normalized graph Laplacian matrix, $\mathbf{L} = \mathbf{I}_n - \mathbf{D}^{-1/2}\mathbf{A}\mathbf{D}^{-1/2}$, is used for the mathematical representation of an undirected graph, where $\mathbf{D}$ is the degree matrix of $\mathbf{A}$. Since the Laplacian is a real symmetric positive semidefinite matrix, we can use the eigen decomposition $\mathbf{L} = \mathbf{U}\mathbf{\Lambda}\mathbf{U}^T$, where $\mathbf{U}\in\mathbb{R}^{n\times n}$ contains eigen vectors and $\mathbf{\Lambda}$ is a diagonal matrix containing eigen values. Thus, with a graph signal, $\mathbf{x}\in\mathbb{R}^{n}$ and a filter $\mathbf{g}\in\mathbb{R}^{n}$ graph convolution is:
\begin{align}
    \mathbf{x}\ast_{\mathcal{G}}\mathbf{g} &= \mathcal{F}^{-1}(\mathcal{F}(\mathbf{x})\odot\mathcal{F}(\mathbf(g)) \\
    &=\mathbf{U}(\mathbf{U}^Tx\odot\mathbf{U}^T\mathbf{g})\\
    &=\mathbf{U}\mathbf{g}_{\theta}\mathbf{U}^T\mathbf{x},
    \label{eq:graph_conv}
\end{align}
where $\odot$ denotes the element-wise product and $\mathbf{g}_{\theta}=\mathrm{diag}(\mathbf{U}^{T}\mathbf{g})$ is the learnable filter. However, evaluating \ref{eq:graph_conv} is costly and note that we first need to perform the eigen decomposition of $\mathbf{L}$, which is also costly operation for large graphs. Thus, \cite{Defferrard2016} parametrizes $\mathbf{g}_{\theta}$ as a truncated expansion, up to order K-1 of Chebyshev polynomials $T_k$. As a result, the graph filtering operation can be written as:
\begin{equation}
    \mathbf{g}_\theta \ast_{\mathcal{G}} \mathbf{x} = \sum_{k=0}^{K-1}\mathbf{\theta}_k T_k(\tilde{L})\mathbf{x},
\end{equation}
where $T_k(\tilde{L})\in\mathbb{R}^{n \times n}$ is the Chebyshev polynomial  of order k evaluated at the scaled Laplacian $\tilde{L} = 2L/\lambda_{\mathrm{max}} - I_n$, and $\mathbf{\theta}\in\mathbb{R}^{K}$ is the vector of Chebyshev coefficients. The $K$ value defines the number of hops we make while aggregating the messages from the neighbors. 

Graph Convolutional Neural networks have homophily and structural equivalence biases, which we employ in crime prediction. We assume that structurally similar nodes will show close behavior. Since graph convolutional networks allow parameter sharing across the nodes in the graph, we can capture the spatial dependencies for different patterns observed in other locations.

To capture the temporal relations we used the graph convolutions with deep memory units proposed in \cite{Seo2018}. Graph Convolutional Gated Recurrent Unit (GCGRU) generalizes the Convolutional Gated Recurrent Unit (ConvGRU) model \cite{Shi2015} to graphs by replacing the Euclidian 2D convolution $\ast$ by the graph convolution $\ast_{\mathcal{G}}$:
\begin{align*}
    z &= \sigma(\mathbf{W}_{xz} \ast_\mathcal{G} \mathbf{x}_t + \mathbf{W}_{hz}\ast_{\mathcal{G}}\mathbf{h}_{t-1}), \\
    r &= \sigma(\mathbf{W}_{xr}\ast_{\mathcal{G}}\mathbf{x}_t + \mathbf{W}_{hr}\ast_{\mathcal{G}}\mathbf{h}_{t-1}), \\
    \tilde{\mathbf{h}} &= \mathrm{tanh}(\mathbf{W}_{xh}\ast_{\mathcal{G}} + \mathbf{W}_{hh}\ast_{\mathcal{G}}(\mathbf{r}\odot\mathbf{h}_{t-1})), \\
    \mathbf{h}_t &= \mathbf{z} \odot \mathbf{h}_{t-1} + (1 - \mathbf{z}) \odot \tilde{\mathbf{h}},
\end{align*}
where $\mathbf{W}_{h.}\in\mathbb{R}^{K\times d_{h} \times d_{h}}$ and $\mathbf{W}_{x.}\in\mathbb{R}^{K\times d_{h} \times d_{x}}$ are the parameters we learn. The hidden dimension, $d_{h}$, the input dimension, $d_{x}$, and $K$ determine number of parameters, which is independent of the number of nodes $N$. The gates $\mathbf{z}$, $\mathbf{r}$, and $\tilde{\mathbf{h}}$ enabling resetting and updating the stored information. 

As show in figure \ref{fig:graph_model}, we stack GCGRU units and use as an encoder to encode the input sequence. At each time slot $t$, each unit takes the previous hidden state, $\mathbf{h}_{t-1}$ and input node feature, $\mathbf{x}_{t}$, to produce the next state $\mathbf{h}_{t}$. Note that we apply these operations for every node on the graph. Thus, in each time step, $t$, we feed the GCGRU unit at each layer with $N$ different node features, $\mathbf{x}_{i,t}$, where the subscript $i$ denotes the node index. Similarly, at each time step, $t$, GCGRU unit outputs the hidden state, $\mathbf{h}_{i, t}$, for each node. As shown in the problem definition, $\mathbf{x}_{i, t}\in\mathbb{R}^{L}$ are the event vectors and $\mathbf{h}_{i,t}\in\mathbb{R}^{d_h}$ are the state vectors that parametrize the likelihood model, $l(x_{i,t}, y_{i,t};\theta(\mathbf{h_{i,t}}))$. 
Each GCGRU output, $\mathbf{h}_{i,t}$, passes to the MLP heads as shown in Figure \ref{fig:graph_model}. Each MLP head is responsible for generating multivariate gaussian parameters shown in equations \ref{eq:mu}-\ref{eq:sigma} for each node:
\begin{align*}
    \mathbf{\mu}_{1} &= \sigma(\mathbf{W}^{T}_{\mu}\mathbf{h}_{1,t} + b_{\mu}), \\
    \mathbf{v}_{1} &= \sigma(\mathbf{W}^{T}_{v}\mathbf{h}_{1,t} + b_{v}), \\
    &\dots \\
    \mathbf{\mu}_{\mathrm{N}} &= \sigma(\mathbf{W}^{T}_{\mu}\mathbf{h}_{\mathrm{N},t} + b_{\mu}), \\
    \mathbf{v}_{\mathrm{N}} &= \sigma(\mathbf{W}^{T}_{v}\mathbf{h}_{\mathrm{N},t} + b_{v}),
\end{align*}
where $\mathbf{W}^{T}_{.}\in\mathbb{R}^{d_h \times 2}$ and $b_{.}$ are the weights we learn, $\sigma(.)$ is the sigmoid activation, and $\mathbf{\mu}_{i}=[\mu_{i, 1}, \mu_{i, 2}]$, $\mathbf{v}_{i}=[v_{i, 1}, v_{i, 2}]$ are the parameter vectors. Since the multivariate distribution is uncorrelated, we write the likelihood model as follows:
\begin{align}
    l(x, y|\mathbf{\mu}, \mathbf{\Sigma}) &= \frac{\mathrm{exp}(\frac{-1}{2} (\mathbf{x}-\mathbf{\mu})^T\mathbf{\Sigma}^{-1}(\mathbf{x} - \mathbf{\mu}))}{2\pi|\mathbf{\Sigma}|^{1/2}} \\
    &= \frac{\mathrm{exp}(\frac{-1}{2}(\frac{(x-\mu_1)^2}{v_1} + \frac{(y-\mu_2)^2}{v_2}))}{2\pi(v_{1}v_{2})^{\frac{1}{2}}},
\end{align}
which allows us to generate likelihoods for any location under the distribution, as we show in the right of figure \ref{fig:graph_model}.

\section{Experiments}
This section compares our model performance with the baseline models on a real and a synthetic dataset. We first describe the datasets with the statistics. Next, we describe each dataset with the results of the experiments. Lastly, we compare the performance of each model.

\subsection{Data Description}

\subsubsection{The Chicago Crime Dataset}
\begin{figure}[t]
\centering
    \begin{subfigure}{0.2\textwidth}
        \centering
        \includegraphics[width=\textwidth]{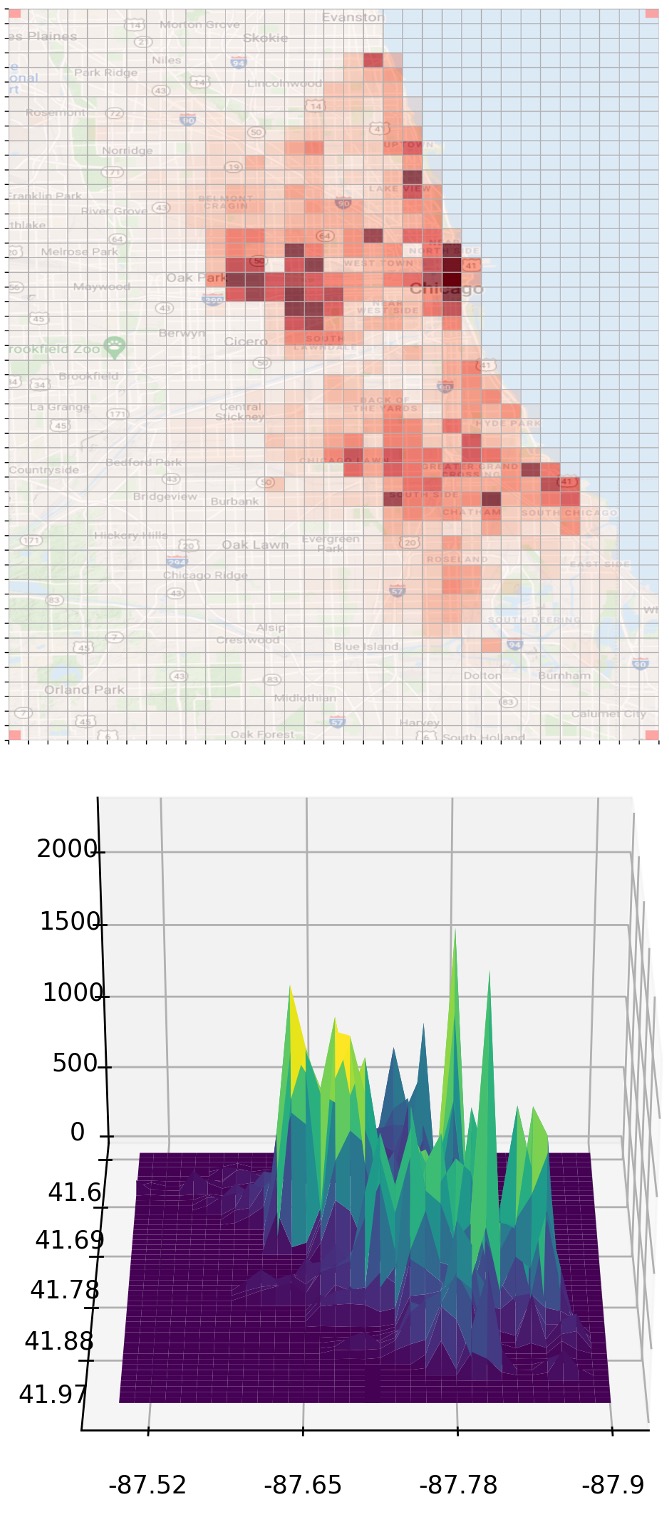}
        \caption{}
        \label{fig:all_dist}
    \end{subfigure}
    \begin{subfigure}{0.22\textwidth}
        \centering
        \includegraphics[width=\textwidth]{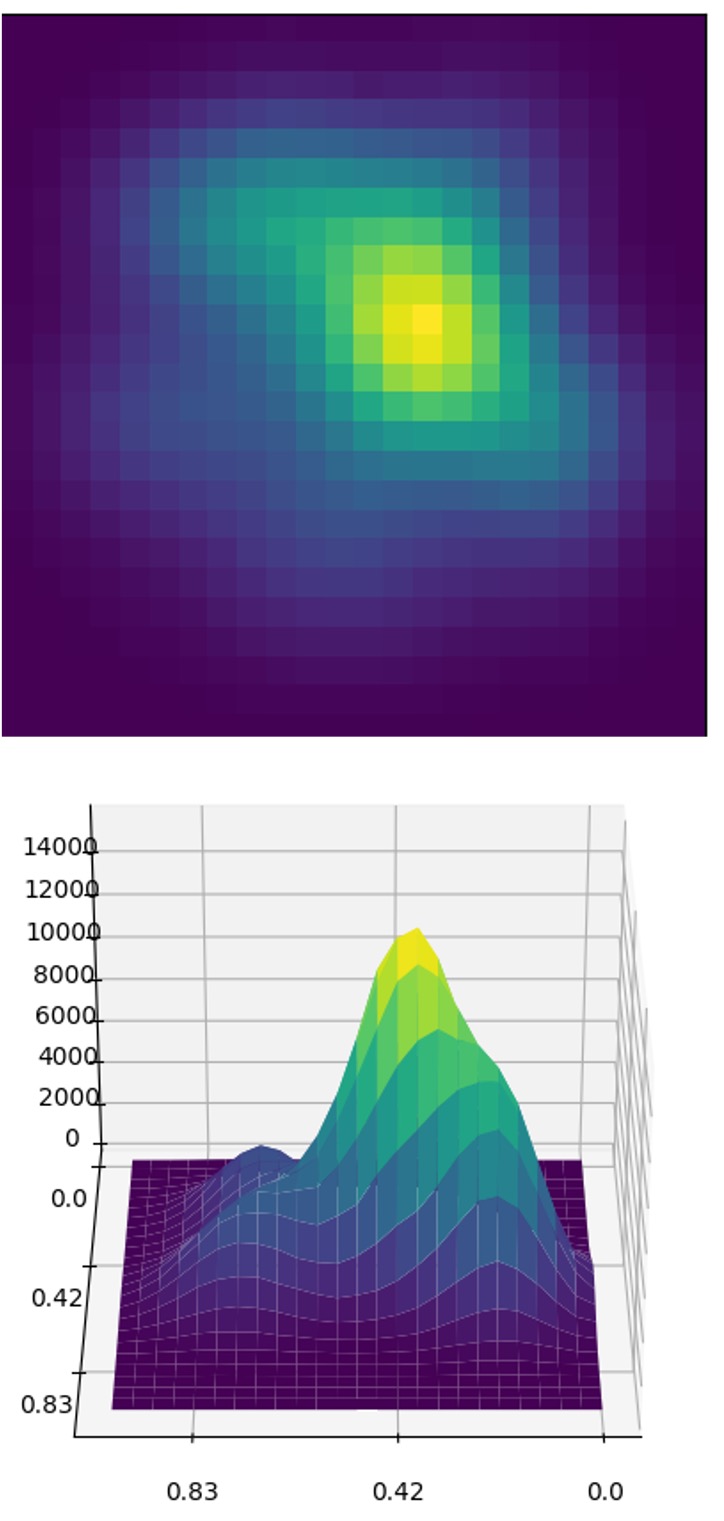}
        \caption{}
        \label{fig:synthetic_dist}
    \end{subfigure}
    \caption{We show the distribution of events for the Chicago crime data (a) and the synthetic data (b).}
    \label{fig:data_dist}
\end{figure}

We collect the Chicago crime data between 2015 and 2019. The working area has a longitude range of $[41.60, 42.05]$ and a latitude range of $[-87.9, -87.5]$, creating a rectangle with $50$km of height and $33$km of width. The rectangle is non-euclidian due to the shape of the Earth, yet, we assumed an euclidian rectangle. We partitioned the region into $50\times 33$ disjointed geographical regions aiming to obtain $1\mathrm{km}\times1\mathrm{km}$ sizes of squares. As we described in the problem definition, we map crime events into an individual geographical region. Figure \ref{fig:all_dist} shows the heat map with the grid shape of the crime distribution regardless of the crime type. We also select the time resolution as $24$ hours. Thus, our target is to predict the locations of any crime event in the partitioned regions for the next $24$ hours. 

Moreover, we perform an exploratory data analysis \footnote{You can find all of our codes to perform analysis and experiments at github.com/sftekin/high-res-crime-forecasting} to observe spatial, temporal, and categorical covariates. Spatially, we concluded that the crime events cluster in particular regions. Temporally, the crime events are self-triggered, and there is a weekly seasonality. Besides, before and during special days such as holidays, weekends we observe similar patterns. Categorically, we perform autocorrelation to features of different crime types to show that crimes correlate with each other. In addition, the crime types; theft, burglary, assault, deceptive practice, criminal damage, narcotics, battery, and robbery forms the $\%90$ percent of the crime data. Thus, we filtered the data by selecting these crime types. 

We perform four experiments where each experiment consists of one year of data. We split the data into ten months of train, one month of validation, and one month of test in each experiment. We report the validation and test scores.

\subsubsection{Synthetic Dataset}
We perform experiments on a synthetic dataset to simulate the performance of our model even further. Our main goal is to create a gaussian distributed spatial data that also has a temporal pattern. In addition, we mimic the crime dataset by creating multiple categories that have self and cross-correlations. 

\begin{figure}[t]
\centering
    \begin{subfigure}{0.35\textwidth}
        \centering
        \includegraphics[width=\textwidth]{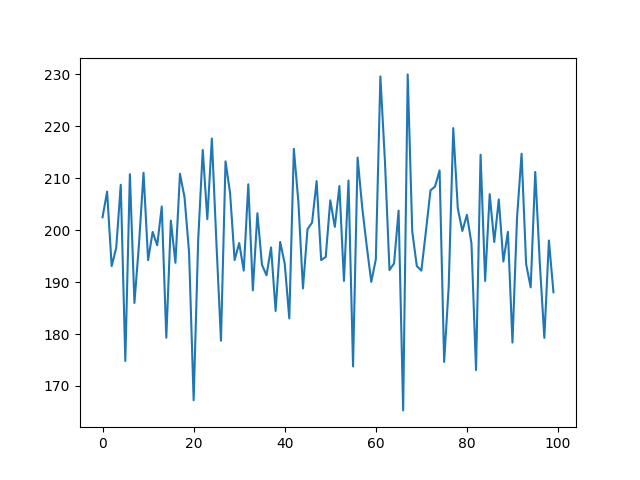}
        \caption{}
        \label{fig:syn_events}
    \end{subfigure}
    \begin{subfigure}{0.35\textwidth}
        \centering
        \includegraphics[width=\textwidth]{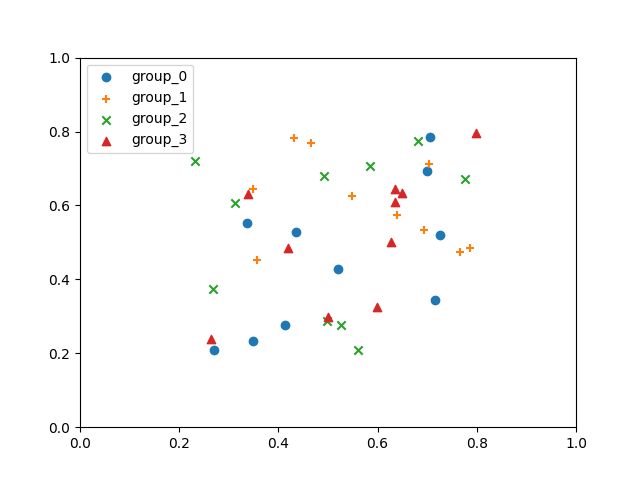}
        \caption{}
        \label{fig:syn_dist}
    \end{subfigure}
    \caption{(a)The first 100 steps of AR series generated for synthetic data. (b) Bi-variate Gaussian Distribution locations for each group}
    \label{fig:syn_creation}
\end{figure}

First, we decided to give a total event count for each time step with a temporal pattern. We model an AR process with $(1, 0)$ parameters, where figure \ref{fig:syn_events} shows the first 100 elements of this series. Second, we set the group count as four and generated ten uniformly distributed density locations between the range of $[0.2, 0.8]$, as shown in figure \ref{fig:syn_dist}. These locations are the mean parameters of the uncorrelated multivariate Gaussian distributions, which variance of them are $0.02, 0.01, 0.005, 0.003$. We sampled from distributions according to the ratios $0.1, 0.2, 0.3, 0.4$. For instance, suppose that the event count for an arbitrary day is $100$, the number of events assigned for groups are $10, 20, 30, 40$, respectively. Third, to create self and cross relations, we create four different profiles and give numbers to each distribution. In the first profile, distributions that have even indices can generate events. The second profile is the opposite of the first profile. For the third profile, indices divisible by four can generate events, and the fourth profile is the opposite of the third. Moreover, after the ten-time step, we switch the behaviors of profiles, where profile one behaves like profile two and vice-versa. Accordingly, we created group relations both temporally and categorically. Figure \ref{fig:synthetic_dist} shows the generated data distribution.

\subsection{HRCF Configurations}

\begin{figure}[t]
    \centering
    \includegraphics[scale=0.09]{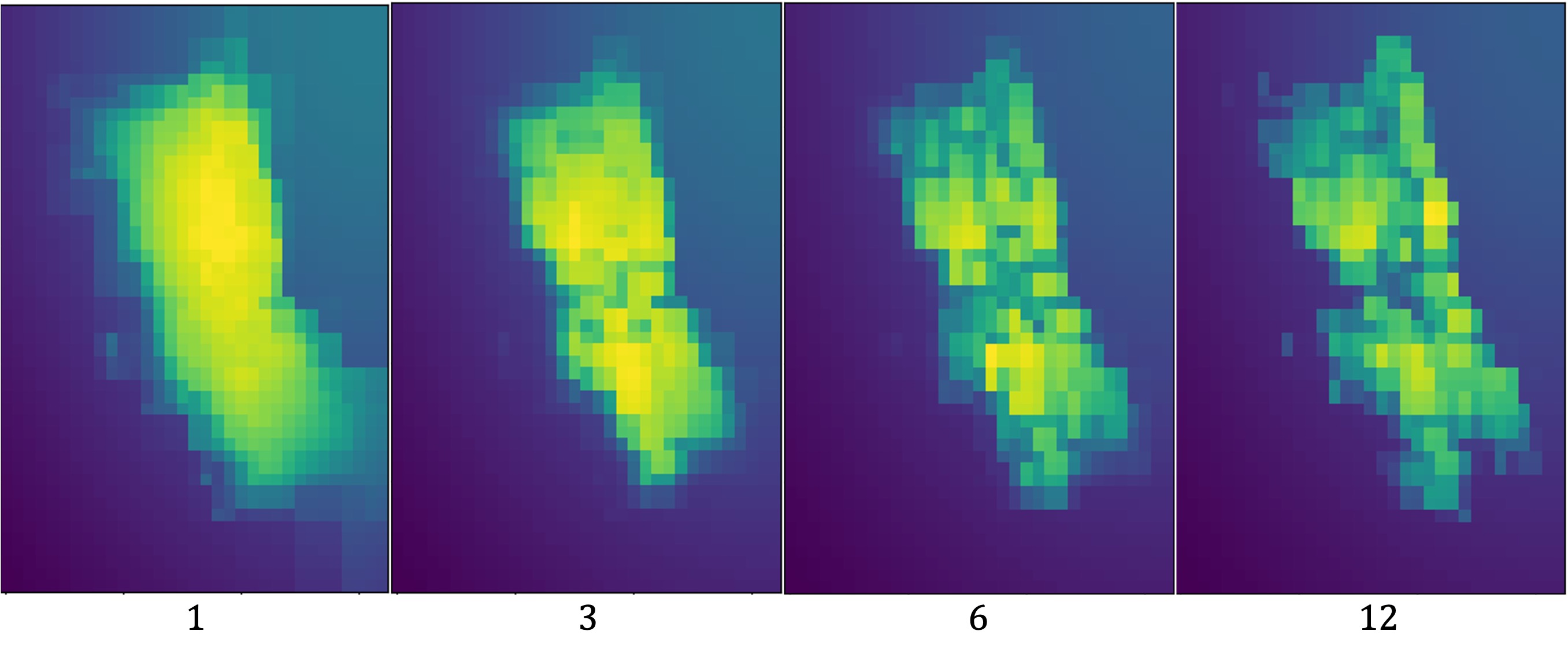}
    \caption{We show the predictions of HRCF model in different epochs.}
    \label{fig:preds}
\end{figure}

\begin{table*}[hbt!]
    \centering
    \begin{tabular}{p{0.8cm} p{0.65cm} p{0.65cm} p{0.65cm} p{0.65cm} p{0.65cm} p{0.65cm} p{0.65cm} p{0.65cm} p{0.65cm} p{0.65cm} p{0.65cm} p{0.65cm}}
        \hline
        \multicolumn{13}{c}{Validation Scores} \\
        \hline
         & \multicolumn{2}{c}{HRCF}  & \multicolumn{2}{c}{ConvLSTM} & \multicolumn{2}{c}{ARIMA} & \multicolumn{2}{c}{RF} & \multicolumn{2}{c}{SVR} & \multicolumn{2}{c}{GPR}\\
        \hline
        Date & F1 & Acc  & F1 & Acc   & F1 & Acc   & F1 & Acc & F1 & Acc & F1 & Acc\\
        \hline
        2015 & \textbf{0.724} & \textbf{0.895} & 0.705 & 0.865 & 0.708 & 0.866 & 0.700 & 0.860 & 0.701 & 0.872 & 0.695 & 0.868\\
        2016 & \textbf{0.728} & \textbf{0.891} & 0.714 & 0.864 & 0.717 & 0.865 & 0.727 & 0.873 & 0.716 & 0.875 & 0.703 & 0.865\\
        2017 & \textbf{0.731} & \textbf{0.895} & 0.709 & 0.865 & 0.716 & 0.868 & 0.713 & 0.865 & 0.718 & 0.879 & 0.703 & 0.869\\
        2018 & \textbf{0.720} & \textbf{0.894} & 0.704 & 0.868 & 0.705 & 0.867 & 0.707 & 0.869 & 0.698 & 0.873 & 0.692 & 0.869\\
        \hline
        \multicolumn{13}{c}{Test Scores} \\
        
        \hline
         & \multicolumn{2}{c}{HRCF}  & \multicolumn{2}{c}{ConvLSTM} & \multicolumn{2}{c}{ARIMA} & \multicolumn{2}{c}{RF} & \multicolumn{2}{c}{SVR} & \multicolumn{2}{c}{GPR}\\
        \hline
        Date & F1 & Acc  & F1 & Acc  & F1 & Acc  & F1 & Acc & F1 & Acc & F1 & Acc\\
        \hline
        2015 & \textbf{0.723} & \textbf{0.888} & 0.713 & 0.871 & 0.705 & 0.863 & 0.718 & 0.874 & 0.709 & 0.876 & 0.694 & 0.867 \\
        2016 & \textbf{0.709} & \textbf{0.881} & 0.698 & 0.863 & 0.718 & 0.876 & 0.693 & 0.855 & 0.697 & 0.870 & 0.684 & 0.864 \\
        2017 & \textbf{0.708} & \textbf{0.877} & 0.699 & 0.865 & 0.686 & 0.853 & 0.697 & 0.862 & 0.697 & 0.872 & 0.686 & 0.866 \\
        2018 & \textbf{0.720} & \textbf{0.884} & 0.708 & 0.870 & 0.712 & 0.869 & 0.716 & 0.872 & 0.708 & 0.876 & 0.693 & 0.869 \\
        \hline
        
    \end{tabular}
    \caption{We show model's validation and test scores on Chicago crime dataset. Bold scores are the best. Date column shows the years the experiment belongs.}
    \label{table:chic_results}
\end{table*}

\begin{table*}[hbt!]
    \centering
    \begin{tabular}{p{0.8cm} p{0.65cm} p{0.65cm} p{0.65cm} p{0.65cm} p{0.65cm} p{0.65cm} p{0.65cm} p{0.65cm} p{0.65cm} p{0.65cm} p{0.65cm} p{0.65cm}} 
        \hline
         & \multicolumn{2}{c}{HRCF}  & \multicolumn{2}{c}{ConvLSTM} & \multicolumn{2}{c}{ARIMA} & \multicolumn{2}{c}{RF} & \multicolumn{2}{c}{SVR} & \multicolumn{2}{c}{GPR}\\
        \hline
        Set & F1 & Acc  & F1 & Acc   & F1 & Acc   & F1 & Acc & F1 & Acc & F1 & Acc\\
        \hline
        Val & \textbf{0.484} & \textbf{0.775} & 0.470 & 0.738 & 0.344 & 0.611 & 0.387 & 0.677 & 0.341 & 0.784 & 0.300 & 0.660 \\
        Test & \textbf{0.491} & \textbf{0.781} & 0.475 & 0.742 & 0.345 & 0.585 & 0.398 & 0.692 & 0.356 & 0.780 & 0.308 & 0.679\\
        \hline
    \end{tabular}
    \caption{We show model's validation and test scores on synthetic dataset. Bold scores are the best.}
    \label{table:syn_results}
\end{table*}

For the subdivision algorithm, we select the threshold value as $1000$ and minimum region size as $(2, 2)$ and obtain 203 regions which is also the number of nodes in our graph. Our model uses $16$ input features total, $8$ for the crime counts for each crime type, $2$ for the locations of nodes, and $6$ for the calendar features such as weekends and holidays. We set the GCGRU layers as three, where hidden dimensions of each layer  $50, 20, 10$. The K value is $3$ in each layer, and we set the bias flag "true" with the symmetric normalization. We used the library of \cite{rozemberczki2021pytorch} to implement GCGRU. The number of layers in MLP is $2$, and the hidden layer contains $64$ nodes. We set the input sequence length as 10 with a batch size of $5$.   

For the training, we use a $0.003$ learning rate, with an early stop tolerance of $5$. We increase the tolerance count by one when the validation loss does not decrease at the end of the epoch.  In every backward pass, we clip the gradients by $10$ to prevent exploding gradient problem. Lastly, we used $0.001$ as L2 loss to regularize the model parameters. 

\subsection{Baseline Models}
We compare our model with the models belonging to three different categories. (i) the conventional time series forecasting methods ARIMA, SVR; (ii) conventional supervised learning algorithm RF and GPR; (iii) recurrent neural network architecture ConvLSTM.

Note that ARIMA, SVR, RF, GPR are regression models; however, we make classification whether a crime event happens or not. Thus, after we make a regression, we select a threshold value for classification. We select the threshold value that maximizes the difference between True Positive Rate (TPR) and  False Positive Rate (FPR). The details of the models are as follows:

\begin{enumerate}
    \item \textbf{Auto-Regressive Integrated Moving Average (ARIMA):}\cite{contreras2003arima} We determine the autoregressive, difference, and moving average parameters according to Akaike Information Criterion (AIC). The parameters that maximize the AIC are $(1,0,1)$. Since the model is only applicable to single time series, the model predicts each driving series of particular cells.
    
    \item \textbf{Support Vector Regression (SVR):} \cite{chang2011libsvm} We choose an RBF kernel and select the other parameters with grid search. We feed the previous ten-time step values of the target series as input features to be fair with the graph model. Similarly, we apply SVR to individual cells and make predictions for each particular cell.
    
    \item \textbf{Random Forest(RF):} \cite{breiman2001random} Similarly, we select the parameters with grid-search and use Mean Squared Error as the criterion. The previous ten-time step values are the input features of the target series, and we apply the model for each particular cell.
    
    \item \textbf{Gaussian Process Regression (GPR):} \cite{rasmussen2006cki} We choose this model to compare with our model better since both models make a gaussian assumption. Model fits spatial data belonging to the former ten time-steps and regresses the following events with 1650 dimensioned multivariate gaussian.
    
    \item \textbf{Convolutional Long-Short Term Memory Units (ConvLSTM):} \cite{Shi2015} We chose this model to compare our graph-based model with a grid-based structured model, taking $50\times33$ shaped input event matrices. We set the input window length as ten and use the encoder-decoder structure as proposed in \cite{Shi2015}. Like in the graph model training, we use early stopping. Furthermore, we train the model with both BCE and MSE loss and take the one with the highest score. 
\end{enumerate}

\subsection{Performance Analysis and Results}

Tables \ref{table:chic_results} and \ref{table:syn_results} show the validation and test scores of models for both datasets. We used the F1 metric, $2\frac{\mathrm{TPR} * \mathrm{FPR}}{\mathrm{TPR} + \mathrm{FPR}}$, where $\mathrm{TPR}$ represents the precision and $\mathrm{FPR}$ represents the recall value. F1 metric measures the overall model performance of how precise and sensitive a model is. Specifically, we also present the accuracy of our models since it is essential in crime prediction. We observe that our model passes the baseline models up to $2-3\%$ for both validation and test scores in both datasets. We perform a statistical significance test with a p-value of $0.05$ comparing baseline results with the HRCF and observe that the HRCF significantly performs better in all metrics. HRCF is not only sensitive to crime events but also accurate. When we look at the score differences between years, we can say that HRCF is more adaptive to the change in the dataset.

In real-life dataset results shown in Table \ref{table:chic_results}, we observe that classical approaches offer comparable performances to deep learning architecture. We provide two reasons for this observation. The first reason is that, as we increase the resolution, we increase the sparsity in the data. Thus, we expect to see similar performance from both classical and deep learning architectures. The second reason is how individual series in particular cells are auto-correlated. In the PACF graph of crime events, we observe that the series have high correlation values in the first ten lags. 

In the synthetic dataset results shown in Table \ref{table:syn_results}, we expect to see higher performance from HRCF since we sampled the data from multiple Gaussian distributions. However, the performance of the GPR model is low, and GPR also assumes the data is coming from a Gaussian distribution. The reason is that the GPR model is not adaptive to temporal changes and can only capture the spatial covariates. From this perspective, we can say that HRCF behaves as a GPR model that can capture both spatial and temporal covariates.

In figure \ref{fig:preds}, we show how the output of HRCF evolves between epochs. The model learns generic bounds of crime locations with minor detail on the high crime rate regions in the first epochs. Then the details on the predictions start to appear in the 6th epoch. After the 12th epoch, the model makes a detailed forecast. However, we observe that model binarizes the output predictions, which means the probability distribution generates likelihoods close to $0$ or $1$. We expect this behavior since we use the BCE loss that makes negative predictions closer to $0$ and positive predictions to $1$. Furthermore, this causes the PDF to assign close values in responsible regions. To assess the generative property of our model on high resolution, we obtain predictions for a $100\times 66$ resolution grid from the HRCF model trained on $50\times 33$. We achieved $36.67$ validation and $36.02$ F1 scores on test sets, which are $1\%$ higher than the ConvLSTM model.

\section{Conclusion}
We studied the problem of high-resolution crime forecasting with a new generative graph neural network architecture, HRCF. We introduced a subdivision algorithm to perform balance sampling on the data to create representative regions. We built our graph according to created regions, parameterized the problem to a likelihood model, and obtained probability density functions belonging to each region. We evaluated HRCF on a real-world and synthetic dataset, and results showed that our model achieves better performance than baseline models. For future work, first, we plan to discover the version with different probability distributions. Second, it is promising to explore dynamic graph creation with a separate graph structure for each time step. Lastly, we applied HRCF on the crime prediction problem; however, it is feasible to apply our model to any spatiotemporal data.



%

\bibliographystyle{IEEEtran}
\bibliography{myref.bib}

\begin{thebibliography}{10}
\providecommand{\url}[1]{#1}
\csname url@samestyle\endcsname
\providecommand{\newblock}{\relax}
\providecommand{\bibinfo}[2]{#2}
\providecommand{\BIBentrySTDinterwordspacing}{\spaceskip=0pt\relax}
\providecommand{\BIBentryALTinterwordstretchfactor}{4}
\providecommand{\BIBentryALTinterwordspacing}{\spaceskip=\fontdimen2\font plus
\BIBentryALTinterwordstretchfactor\fontdimen3\font minus
  \fontdimen4\font\relax}
\providecommand{\BIBforeignlanguage}[2]{{%
\expandafter\ifx\csname l@#1\endcsname\relax
\typeout{** WARNING: IEEEtran.bst: No hyphenation pattern has been}%
\typeout{** loaded for the language `#1'. Using the pattern for}%
\typeout{** the default language instead.}%
\else
\language=\csname l@#1\endcsname
\fi
#2}}
\providecommand{\BIBdecl}{\relax}
\BIBdecl

\bibitem{ehrlich1975relation}
I.~Ehrlich, ``On the relation between education and crime,'' National Bureau of
  Economic Research, Tech. Rep., 1975.

\bibitem{braithwaite1989crime}
J.~Braithwaite \emph{et~al.}, \emph{Crime, shame and reintegration}.\hskip 1em
  plus 0.5em minus 0.4em\relax Cambridge University Press, 1989.

\bibitem{patterson1991poverty}
E.~B. Patterson, ``Poverty, income inequality, and community crime rates,''
  \emph{Criminology}, vol.~29, no.~4, pp. 755--776, 1991.

\bibitem{freeman1999economics}
R.~B. Freeman, ``The economics of crime,'' \emph{Handbook of labor economics},
  vol.~3, pp. 3529--3571, 1999.

\bibitem{Toole2011}
J.~L. Toole, N.~Eagle, and J.~B. Plotkin, ``{Spatiotemporal correlations in
  criminal offense records},'' \emph{ACM Transactions on Intelligent Systems
  and Technology}, vol.~2, no.~4, 2011.

\bibitem{wang2012automatic}
X.~Wang, M.~S. Gerber, and D.~E. Brown, ``Automatic crime prediction using
  events extracted from twitter posts,'' in \emph{International conference on
  social computing, behavioral-cultural modeling, and prediction}.\hskip 1em
  plus 0.5em minus 0.4em\relax Springer, 2012, pp. 231--238.

\bibitem{gerber2014predicting}
M.~S. Gerber, ``Predicting crime using twitter and kernel density estimation,''
  \emph{Decision Support Systems}, vol.~61, pp. 115--125, 2014.

\bibitem{Wang2016}
H.~Wang, D.~Kifer, C.~Graif, and Z.~Li, ``{Crime rate inference with big
  data},'' \emph{Proceedings of the ACM SIGKDD International Conference on
  Knowledge Discovery and Data Mining}, vol. 13-17-Augu, pp. 635--644, 2016.

\bibitem{Bowers2004}
K.~J. Bowers, S.~D. Johnson, and K.~Pease, ``{Prospective hot-spotting: The
  future of crime mapping?}'' \emph{British Journal of Criminology}, vol.~44,
  no.~5, pp. 641--658, 2004.

\bibitem{Chainey2008}
S.~Chainey, L.~Tompson, and S.~Uhlig, ``{The Utility of Hotspot Mapping for
  Predicting Spatial Patterns of Crime},'' \emph{Security Journal}, vol.~21,
  no. 1-2, pp. 4--28, 2008.

\bibitem{Mohler2011}
G.~O. Mohler, M.~B. Short, P.~J. Brantingham, F.~P. Schoenberg, and G.~E. Tita,
  ``{Self-exciting point process modeling of crime},'' \emph{Journal of the
  American Statistical Association}, vol. 106, no. 493, pp. 100--108, 2011.

\bibitem{Mohler2014}
\BIBentryALTinterwordspacing
G.~Mohler, ``{Marked point process hotspot maps for homicide and gun crime
  prediction in Chicago},'' \emph{International Journal of Forecasting},
  vol.~30, no.~3, pp. 491--497, 2014. [Online]. Available:
  \url{http://dx.doi.org/10.1016/j.ijforecast.2014.01.004}
\BIBentrySTDinterwordspacing

\bibitem{Du2016}
N.~Du, H.~Dai, R.~Trivedi, U.~Upadhyay, M.~Gomez-Rodriguez, and L.~Song,
  ``{Recurrent Marked Temporal Point Processes},'' pp. 1555--1564, 2016.

\bibitem{Zhang2016}
J.~Zhang, Y.~Zheng, D.~Qi, R.~Li, and X.~Yi, ``{DNN-Based Prediction Model for
  Spatio-Temporal Data},'' 2016.

\bibitem{Zhang2017}
J.~Zhang, Y.~Zheng, and D.~Qi, ``{Deep spatio-temporal residual networks for
  citywide crowd flows prediction},'' \emph{31st AAAI Conference on Artificial
  Intelligence, AAAI 2017}, pp. 1655--1661, 2017.

\bibitem{Wang2017}
\BIBentryALTinterwordspacing
B.~Wang, D.~Zhang, D.~Zhang, P.~J. Brantingham, and A.~L. Bertozzi, ``{Deep
  Learning for Real Time Crime Forecasting},'' pp. 33--36, 2017. [Online].
  Available: \url{http://arxiv.org/abs/1707.03340}
\BIBentrySTDinterwordspacing

\bibitem{Huang2019}
C.~Huang, X.~Wu, C.~Zhang, D.~Yin, J.~Zhao, and N.~V. Chawla, ``{MIST: A
  multiview and multimodal spatial-temporal learning framework for citywide
  abnormal event forecasting},'' \emph{The Web Conference 2019 - Proceedings of
  the World Wide Web Conference, WWW 2019}, pp. 717--728, 2019.

\bibitem{wu2020comprehensive}
Z.~Wu, S.~Pan, F.~Chen, G.~Long, C.~Zhang, and S.~Y. Philip, ``A comprehensive
  survey on graph neural networks,'' \emph{IEEE transactions on neural networks
  and learning systems}, vol.~32, no.~1, pp. 4--24, 2020.

\bibitem{Jain2016}
A.~Jain, A.~R. Zamir, S.~Savarese, and A.~Saxena, ``{Structural-RNN: Deep
  learning on spatio-temporal graphs},'' \emph{Proceedings of the IEEE Computer
  Society Conference on Computer Vision and Pattern Recognition}, vol.
  2016-Decem, pp. 5308--5317, 2016.

\bibitem{Wang2018}
\BIBentryALTinterwordspacing
B.~Wang, X.~Luo, F.~Zhang, B.~Yuan, A.~L. Bertozzi, and P.~J. Brantingham,
  ``{Graph-Based Deep Modeling and Real Time Forecasting of Sparse
  Spatio-Temporal Data},'' 2018. [Online]. Available:
  \url{http://arxiv.org/abs/1804.00684}
\BIBentrySTDinterwordspacing

\bibitem{Sun2021}
J.~Sun, M.~Yue, Z.~Lin, X.~Yang, L.~Nocera, G.~Kahn, and C.~Shahabi,
  ``{CrimeForecaster: Crime Prediction by Exploiting the Geographical
  Neighborhoods' Spatiotemporal Dependencies},'' \emph{Lecture Notes in
  Computer Science (including subseries Lecture Notes in Artificial
  Intelligence and Lecture Notes in Bioinformatics)}, vol. 12461 LNAI, pp.
  52--67, 2021.

\bibitem{Wang2021}
\BIBentryALTinterwordspacing
C.~Wang, Z.~Lin, X.~Yang, J.~Sun, M.~Yue, and C.~Shahabi, ``{HAGEN:
  Homophily-Aware Graph Convolutional Recurrent Network for Crime
  Forecasting},'' 2021. [Online]. Available:
  \url{http://arxiv.org/abs/2109.12846}
\BIBentrySTDinterwordspacing

\bibitem{Defferrard2016}
M.~Defferrard, X.~Bresson, and P.~Vandergheynst, ``{Convolutional neural
  networks on graphs with fast localized spectral filtering},'' \emph{Advances
  in Neural Information Processing Systems}, no. Nips, pp. 3844--3852, 2016.

\bibitem{Seo2018}
Y.~Seo, M.~Defferrard, P.~Vandergheynst, and X.~Bresson, ``{Structured sequence
  modeling with graph convolutional recurrent networks},'' \emph{Lecture Notes
  in Computer Science (including subseries Lecture Notes in Artificial
  Intelligence and Lecture Notes in Bioinformatics)}, vol. 11301 LNCS, no.
  2013, pp. 362--373, 2018.

\bibitem{Shi2015}
X.~Shi, Z.~Chen, H.~Wang, D.~Y. Yeung, W.~K. Wong, and W.~C. Woo,
  ``{Convolutional LSTM network: A machine learning approach for precipitation
  nowcasting},'' \emph{Advances in Neural Information Processing Systems}, vol.
  2015-Janua, pp. 802--810, 2015.

\bibitem{rozemberczki2021pytorch}
B.~Rozemberczki, P.~Scherer, Y.~He, G.~Panagopoulos, A.~Riedel, M.~Astefanoaei,
  O.~Kiss, F.~Beres, , G.~Lopez, N.~Collignon, and R.~Sarkar, ``{PyTorch
  Geometric Temporal: Spatiotemporal Signal Processing with Neural Machine
  Learning Models},'' in \emph{Proceedings of the 30th ACM International
  Conference on Information and Knowledge Management}, 2021, p. 4564–4573.

\bibitem{contreras2003arima}
J.~Contreras, R.~Espinola, F.~J. Nogales, and A.~J. Conejo, ``Arima models to
  predict next-day electricity prices,'' \emph{IEEE transactions on power
  systems}, vol.~18, no.~3, pp. 1014--1020, 2003.

\bibitem{chang2011libsvm}
C.-C. Chang and C.-J. Lin, ``Libsvm: a library for support vector machines,''
  \emph{ACM transactions on intelligent systems and technology (TIST)}, vol.~2,
  no.~3, pp. 1--27, 2011.

\bibitem{breiman2001random}
L.~Breiman, ``Random forests,'' \emph{Machine learning}, vol.~45, no.~1, pp.
  5--32, 2001.

\bibitem{rasmussen2006cki}
C.~Rasmussen, ``Cki williams gaussian processes for machine learning,'' 2006.

\end{thebibliography}

%

\end{document}